\documentclass{article}

\usepackage{ijcai23}

\usepackage{times}
\usepackage{soul}
\usepackage{url}
\usepackage[hidelinks]{hyperref}
\usepackage[utf8]{inputenc}
\usepackage[small]{caption}
\usepackage{graphicx}
\usepackage{amsmath}
\usepackage{amsthm}
\usepackage{booktabs}
\usepackage{algorithm}
\usepackage{algorithmic}
\usepackage[switch]{lineno}

\usepackage{algorithm}
\usepackage{algorithmic}
\usepackage{amssymb}
\usepackage{multirow}
\linenumbers

\title{ContrastMotion: Self-supervised Scene Motion Learning for Large-Scale LiDAR Point Clouds}

\begin{document}

\maketitle
\section{Reviewer 1}
\subsection{For 2D pillar flow to 3D point flow, is the motion in z-axis omitted?}
Response: When scattering 2D pillar flow 3D point flow, the motion in z-axis is set to 0. Please note that the evaluation metrics we use, such as EPE3D, Acc3DS, Acc3DR, etc. still evaluate Z-axis predictions.

\subsection{'Ground Mask' in Table 4. The Ground Mask is used in other experiments (other tables)? As shown in Tab. 4, when using Ground Mask, PointInfoNCE Loss performs not well in comparison with the proposed SD-loss. While, the authors do not provide the comparisons when removing the Ground Mask. It weird. }

\begin{table*}
    \centering
        \caption{\textbf{Ablation study for different modules of ContrastMotion.} We evaluate each model on nuScenes motion prediction and report mean errors. Ground Mask: Removing estimated ground points before feeding to network. Gated Block: \textit{Gated Multi-frame Fusion Block}.}
    \scalebox{0.9}{
    \begin{tabular}{c|c c|c|c|c|c|c}
    \hline
        Method & SD--Loss & PointInfoNCE Loss & Ground Mask & GMF & Static & Speed $\le$ 5m/s & Speed $>$ 5m/s\\
        \hline
         (newcomer) & & \checkmark & & & 0.1396 & 0.9147 & 4.2056 \\
         (a) & \checkmark & & & & 0.1377 & 0.7791 & 3.9497 \\
         (b) & \checkmark & & & \checkmark & \textbf{0.0682} & 0.4579 & 3.9872 \\
         (c) & \checkmark & & \checkmark & \checkmark & 0.0829 & \textbf{0.4522} & \textbf{3.5266} \\
         (d) & & \checkmark & \checkmark & \checkmark & 0.0837 & 0.6546 & 3.9601 \\
    \hline
    \end{tabular}
    }
    \label{tab:ABL}
\end{table*}

Response: Actually, the Ground Mask is used in ablation experiments, but due to page limitations, we put the content in the Supplementary Material (Section 2.1 Ground Removal, Section 2.2 Ablation Experiments of Ground Points) where we describe how to remove ground points using an unsupervised approach and the effect of ground points on the model. In fact, the results of the individual comparisons of SD-Loss and PointInfoNCE are similar to those of method (c) and (d). We now increase the comparison items (newcomer) as shown in Table~\ref{tab:ABL}, we find that model (a) achieves better performances in all the groups compared to model (newcomer). And, the mean errors difference in static group between (a) and (newcomer) is the smallest relative to the movable groups, which is also observed in the comparison of (c) and (d). We believe the reason is that movable objects are more likely to appear the case of correspondence points missing which lead to higher mean errors difference in movable groups between (a) and (newcomer). We will include comparisons (newcomer) and related analysis in the final version.

\section{Reviewer 2}
\subsection{The LiDAR frames of T and T+1 in your implementation are orginated from the same point cloud by applying data augmentation techniques to ensure pairwise correspondence labels. Such a contrastive learning design is similar to the unsupervised pertaining methods like PointContrast, SegContrast, GCC-3D, ProposalContrast, and so on. Although you mentioned in the related work that PointContrast only considers point-level correspondence, follow-up works like SegContrast, GCC-3D and ProposalContrast can achieve pseudo object-level correspondence. Have you tried the performances of these methods on the scene flow prediction task by substituting the contrastive discrimination parts of your method.}

Response: As we known, SegContrast and DepthContrast are proposed for pretraining, GCC-3D and ProposalContrast are proposed for object detection, and all of them have a contrastive learning design. However, There are limitations when these contrast learning designs are used for scene motion task. DepthContrast only involves point-level correspondence matching which suffer from the case of correspondence points missing in real scenes. SegContrast, GCC-3D and ProposalContrast achieve pseudo object-level correspondence which is a coarse-grained contrast compared to our pillar-level. Object-level correspondence predicts the same motion for each part of the object and thus may lead to inaccurate motion prediction, especially when the object is rotating (e.g. a car turning) and the motion of each part of the object is different. However, Object-level correspondence from these works can be leveraged to enhance the entirety and consistency of object, and future work will build on this for more exploration.

\subsection{What is the effect of granularity of pillar features? You assign the same scene flow predictions to all points within the same pillar, while points within a pillar generally belong to different moving or static subjects. Thus a pillar with large pillar size can lead to inaccurate assignment. It is susceptible whether such an assignment is qualified to handle the dense prediction task like scene flow prediction.}
Response: In fact, the granularity of the pillar (which we call pillar size) has a significant impact on the performance of the model, and a pillar with large pillar size can lead to inaccurate assignment. However, compared with point-based, the advantages of pillar-based are reflected in large-scale point clouds processing and inference speed, which is one of the core points of our paper (as also PillarMotion). The pillar size should be set carefully to balance accuracy and efficiency, and in subsection Implementation Details, we show our pillar size [0.25m, 0.25m].

PillarMotion: Self-Supervised Pillar Motion Learning for Autonomous Driving (CVPR 2021).
\subsection{Compared state-of-the-art methods are old. Comparison with newer methods like RigidFlow (CVPR 2022) should be given.}

\begin{table}
    \centering
        \caption{\textbf{Evaluation results on FT3D $\rightarrow$ KITTI Scene Flow.} $^\ddag$: 8192 points (downsampling) for training and test. $^\dag$: All points for training and test.}
    \scalebox{0.7}{
    \begin{tabular}{c|c|c|c|c}
    \hline
      Method & EPE3D(m)$\downarrow$ & Acc3DS$\uparrow$ & Acc3DR$\uparrow$ & Outliers3D$\downarrow$ \\
    \hline
     RigidFlow$^\dag$ & \textbf{0.0619} & 0.7237 & 0.8923 & 0.2618 \\
     ContrastMotion (ours)$^\dag$  & 0.0656 & \textbf{0.7382} & \textbf{0.9417} & \textbf{0.1847} \\
    \hline
    \end{tabular}
    }
    \label{tab:KITTT_RESULTS}
\end{table}

Response: As shown in Table\ref{tab:KITTT_RESULTS}, our ContrastMotion is only slightly weaker in terms of EPE3D compared with RigidFlow. The reason why there is no comparison with RigidFlow in origin paper is that we cannot obtain its performance in the non-downsampled case. And we focus on large-scale point cloud scene motion learning (our title), so we temporarily have no experiments conducted on downsampled point clouds.

\section{Reviewer 3}
\subsection{I am still suspicious about the functionality of data augmentation. Intuitively, augmentation is a larger superset of the actual LiDAR movement, where the foreground objects would rotate, translate. etc. However, the data augmentation for all points in the scene would result in the same motion for all the objects. Is this useful for actual motion prediction of the foreground objects?}
Response:  We quite understand your doubts. It’s true that the data augmentation for all points in the scene would result in the same motion for all the objects, so it seems that the data augmentation only works for point cloud registration which estimates the transformation matrix between point cloud pairs, while scene motion estimates fine-grained motion and the motion of the foreground objects is different. The reason why ContrastMotion is working relies on the fact that we use a pillar association process instead of the conventional prediction head. Due to the presence of pillar association, the model predicts pillar correspondence probabilities and is agnostic to pillar motion. So whether the foreground objects have the same motion (data augmentation samples) or not (real point cloud pairs), the model correctly predicts pillar correspondence probabilities, and then we correctly calculate scene motion through pillar correspondence probabilities. In 3.6 subsection Analysis, we also analyzed this problem.

\subsection{The compression with state-of-the-art methods in CVPR2022 and 2023 are missing.}
Response: As shown in Table\ref{tab:KITTT_RESULTS}, our ContrastMotion is only slightly weaker in terms of EPE3D compared with RigidFlow (CVPR 2022). The reason why there is no comparison with RigidFlow in origin paper is that we cannot obtain its performance in the non-downsampled case. And we focus on large-scale point cloud scene motion learning (our title), so we temporarily have no experiments conducted on downsampled point clouds.

\subsection{How will your method perform when you apply the pretrained weights to the fine-tuning setting with fully training set?}
Response: It's a great idea, but time for rebuttal is a little tight for us, and we'll try it next.
\section{Reviewer 4}
\subsection{One minor question: The proposed method is about 5x slower than some of the existing methods compared for inference speed, mentioning that using parallel computing would boost the speed. Does that mean other methods are benefitted from parallel processing?}
Response:  As you seen in Section Runtime Analysis of Supplementary Material, the pillar association takes 95ms (80\% of total cost). We want to use parallel computing to improve the inference speed of pillar association process. Other methods already benefit from parallel processing due to the use of prediction heads.
\subsection{I hope the code to be available no later than the conference date.}
Response:  Thank you for your approval of our work. Our code is built on PillarMotion's open source code and has only a few closing works left. We release the code as soon as possible and no later than June.

\end{document}